\documentclass[journal,twoside,web]{ieeecolor}
\usepackage{tmi}
\usepackage{cite}
\usepackage{amsmath,amssymb,amsfonts}
\usepackage{algorithmic}
\usepackage{graphicx}
\usepackage{textcomp}

\usepackage{multirow} 
\usepackage{makecell} 
\usepackage{booktabs}
\usepackage{bbm}
\usepackage{mathtools}
\usepackage{bbold}
\usepackage{caption}
\usepackage{bm}
\usepackage{marvosym}
\usepackage[colorlinks,linkcolor=blue]{hyperref}

\newcommand{\ie}{\textit{i.e.}, }
\newcommand{\eg}{\textit{e.g.}, }
\newcommand{\etal}{\textit{et al. }}

\newcommand{\wrt}{\textit{w.r.t. }}

\def\BibTeX{{\rm B\kern-.05em{\sc i\kern-.025em b}\kern-.08em
    T\kern-.1667em\lower.7ex\hbox{E}\kern-.125emX}}
\begin{document}
\title{Generating and Weighting Semantically Consistent Sample Pairs for Ultrasound Contrastive Learning}
\author{Yixiong Chen, Chunhui Zhang, Chris H.Q. Ding, and Li Liu~\IEEEmembership{Member,~IEEE,}
\thanks{This work is supported by the National Natural Science Foundation of China (No. 62101351), the Guangdong Basic and Applied Basic Research Foundation (No. 2020A1515110376), Shenzhen Outstanding Scientific and Technological Innovation Talents Ph.D. Startup Project (No. RCBS20210609104447108), the Key-Area Research and Development Program of Guangdong Province (No. 2020B0101350001).}
\thanks{Yixiong Chen is with the School of Data Science, the Chinese University of Hong Kong (Shenzhen), 518172 Shenzhen, China (e-mail: yixiongchen@link.cuhk.edu.cn).}
\thanks{Chunhui Zhang is with the Institute of Information Engineering, Chinese Academy of Sciences, 100049 Beijing, China (e-mail: zhangchunhui17@mails.ucas.ac.cn).}
\thanks{Chris H.Q. Ding is with the School of Data Science, the Chinese University of Hong Kong (Shenzhen), 518172 Shenzhen, China (e-mail: chrisding@cuhk.edu.cn).}
\thanks{Li Liu is with the Shenzhen Research Institute of Big Data, the Chinese University of Hong Kong (Shenzhen), 518172 Shenzhen, China (corresponding author, e-mail: liuli@cuhk.edu.cn).}
}
\maketitle

\begin{abstract}
Well-annotated medical datasets enable deep neural networks (DNNs) to gain strong power in extracting lesion-related features. Building such large and well-designed medical datasets is costly due to the need for high-level expertise. Model pre-training based on ImageNet is a common practice to gain better generalization when the data amount is limited. However, it suffers from the domain gap between natural and medical images. In this work, we pre-train DNNs on ultrasound (US) domains instead of ImageNet to reduce the domain gap in medical US applications. To learn US image representations based on unlabeled US videos, we propose a novel meta-learning-based contrastive learning method, namely Meta Ultrasound Contrastive Learning (Meta-USCL). To tackle the key challenge of obtaining semantically consistent sample pairs for contrastive learning, we present a positive pair generation module along with an automatic sample weighting module based on meta-learning. Experimental results on multiple computer-aided diagnosis (CAD) problems, including pneumonia detection, breast cancer classification, and breast tumor segmentation, show that the proposed self-supervised method reaches state-of-the-art (SOTA). The codes are available at \href{https://github.com/Schuture/Meta-USCL}{https://github.com/Schuture/Meta-USCL}.
\end{abstract}


\section{Introduction}
\label{sec:introduction}

Ultrasound (US) medical imaging system plays a vital role in computer-aided diagnosis (CAD) for its affordable cost~\cite{born2020pocovid} and being physically portable~\cite{thomenius2009miniaturization}. 
In previous literature, remarkable progress has been made in US image analysis~\cite{singal2009meta,chi2017thyroid}, primarily due to the increasingly powerful deep neural network (DNN) architectures and the growing amount of annotated medical datasets. However, most medical applications still lack sufficient data, resulting in the overfitting of complex DNNs~\cite{tajbakhsh2016convolutional}. Transfer learning~\cite{sharif2014cnn}, consisting of pre-training and fine-tuning, is an effective method to mitigate this problem. In the pre-training stage, DNNs learn generic visual representations from large datasets like ImageNet~\cite{deng2009imagenet}. This process is called \textit{representation learning}. In the fine-tuning stage, models adapt to the downstream tasks by reusing learned knowledge~\cite{matsoukas2022makes} and slightly updating pre-trained parameters. ImageNet contains millions of natural images, which makes it a widely-used pre-training dataset to learn abundant visual representations. Though effective, there is one problem that ImageNet pre-training usually meets when applied for medical applications, namely domain gap. Visual representations from the natural image domain of ImageNet would not necessarily be transferable to the downstream medical domains because they are quite dissimilar. The large domain gap between the source and target data will likely reduce the performance gain of transfer learning~\cite{tajbakhsh2016convolutional}.

To narrow the domain gaps between pre-training and downstream medical tasks, many medical representation learning methods~\cite{ning2021deep,Jianbo2020MICCAI,nguyen2020self,zhou2019models,zhou2020comparing} were proposed. They learn representations directly from the same medical domains as the downstream tasks. With the recent development of self-supervised learning~\cite{Ting2020CVPR,chen2020big,he2020momentum,ghesu2022self}, DNNs are able to exploit the existing unlabeled data from various sources to learn powerful representations. Medical models trained with self-supervised learning are shown to be highly transferable on a wide range of medical modalities like CT, MRI, and X-ray~\cite{nguyen2020self,zhou2019models,zhou2020comparing,sowrirajan2021moco}. There are also self-supervised methods proposed for the US, but they either require clear anatomical structures~\cite{jiao2020self} or aligned video-speech correspondence~\cite{Jianbo2020MICCAI} to train, which restricts their applications. A method that can be applied to any regular free-hand US videos~\cite{baumgartner2017sononet} is needed.

\begin{figure*}[t]
\centering\centerline{\includegraphics[width=1.0\linewidth]{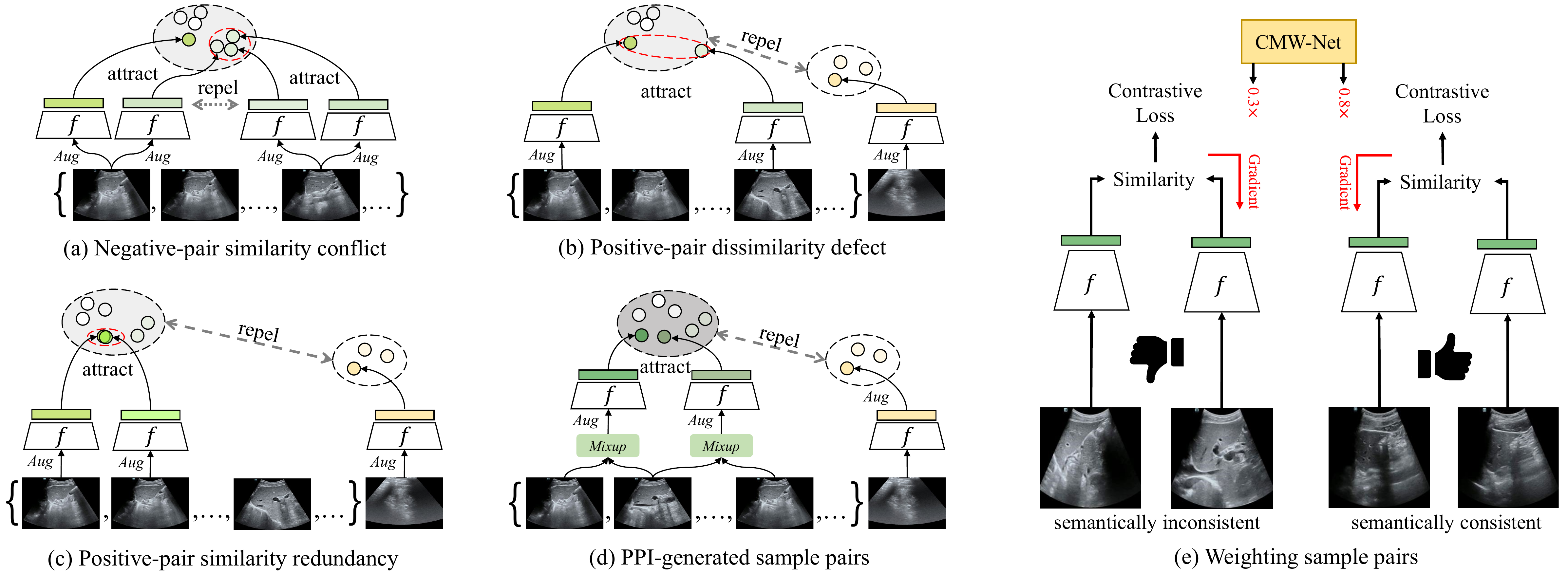}}
\caption{The motivation of Meta-USCL. ``Aug" indicates the random data augmentation. (a) Negative-pair similarity conflict: two positive pairs come from the same video, pushing them apart confuses the training. (b) Positive-pair dissimilarity defect: samples in a positive pair are semantically different. (c) Positive-pair similarity redundancy: samples in a positive pair are near identical. (d) PPI-generated positive pairs have moderate appearance discrepancies and small semantic distances. (e) CMW-Net makes the training focus more on those semantically consistent sample pairs.}
\label{fig:Motivation}
\end{figure*}

In this work, we regard contrastive learning~\cite{Ting2020CVPR,he2020momentum} as an ideal self-supervised learning baseline for US applications, because it shows promising performance in learning robust visual representations comparable to the supervised counterparts. The idea of contrastive learning is making two semantically similar samples (\ie a positive sample pair) attract each other in the feature space while two dissimilar samples (\ie a negative sample pair) repel each other. This work tries to utilize this idea, and tailor a contrastive learning algorithm to learn robust US representations from unlabeled videos. However, most previous contrastive methods (\eg SimCLR~\cite{Ting2020CVPR}, MoCo~\cite{he2020momentum}) were proposed for image data, and they usually use an image to augment twice as a positive sample pair while two samples are generated from different images as a negative pair. Sample pairs generated in this way can hardly work well for US videos. We summarize three main problems of adapting contrastive learning for US videos: 1) \textit{Negative-pair similarity conflict} (Fig. \ref{fig:Motivation} (a)) appears when multiple positive pairs are generated from the same video. In this case, typical contrastive learning regards two samples generated from two frames as a negative pair, and their similarity may be higher than that between two randomly augmented positive samples. Pushing two negative samples with high semantic consistency apart misleads the training. It can be avoided by only generating one positive pair from a video. Moreover, considering a short US video clip usually contains the same lesion from the same patient, it is natural to take two frames from the same video as positive pair because they probably lead to similar diagnosis results. This improvement exploits the slight dissimilarity between frames as the natural perturbation, reaching the same goal as image augmentation, and enriches the feature diversity involved in a positive pair. 2) \textit{Positive-pair dissimilarity defect} (Fig. \ref{fig:Motivation} (b)) makes a positive pair no longer consistent in semantics, which is usually caused by too intense image transformation or significant frame differences. 3) \textit{Positive-pair similarity redundancy} (Fig. \ref{fig:Motivation} (c)) appears in the opposite case. When the random image transformation is too weak, or the frames only differ slightly, the appearance change of a positive pair teaches the DNN nothing because the model can obtain close features even if it does not have visual discriminative ability. The theory of contrastive learning~\cite{tian2020makes} claims that what should be contrasted in contrastive learning should be the detailed appearance differences between a positive pair that do not affect their semantic meaning in common.

To tackle the positive-pair dissimilarity defect and similarity redundancy problems, we propose a \textbf{P}ositive \textbf{P}air \textbf{I}nterpolation (PPI) module to generate positive pairs by interpolating frames from a video. Two image samples generated from PPI share common information of an ``anchor" frame to remain semantically consistent, and randomly mix different appearances from the other two frames. The coefficients for mixing three frames are randomly sampled.
This method reduces the risk of generating positive pairs that are too dissimilar (positive-pair dissimilarity defect) or similar (positive-pair similarity redundancy). It also enriches the feature of positive pairs by the multi-frame synthesis.
A PPI-generated positive pair acts as two points with moderate distance in a semantic cluster (Fig. \ref{fig:Motivation} (d)). 

\begin{figure*}[t]
\centering
\includegraphics[width=0.85\linewidth]{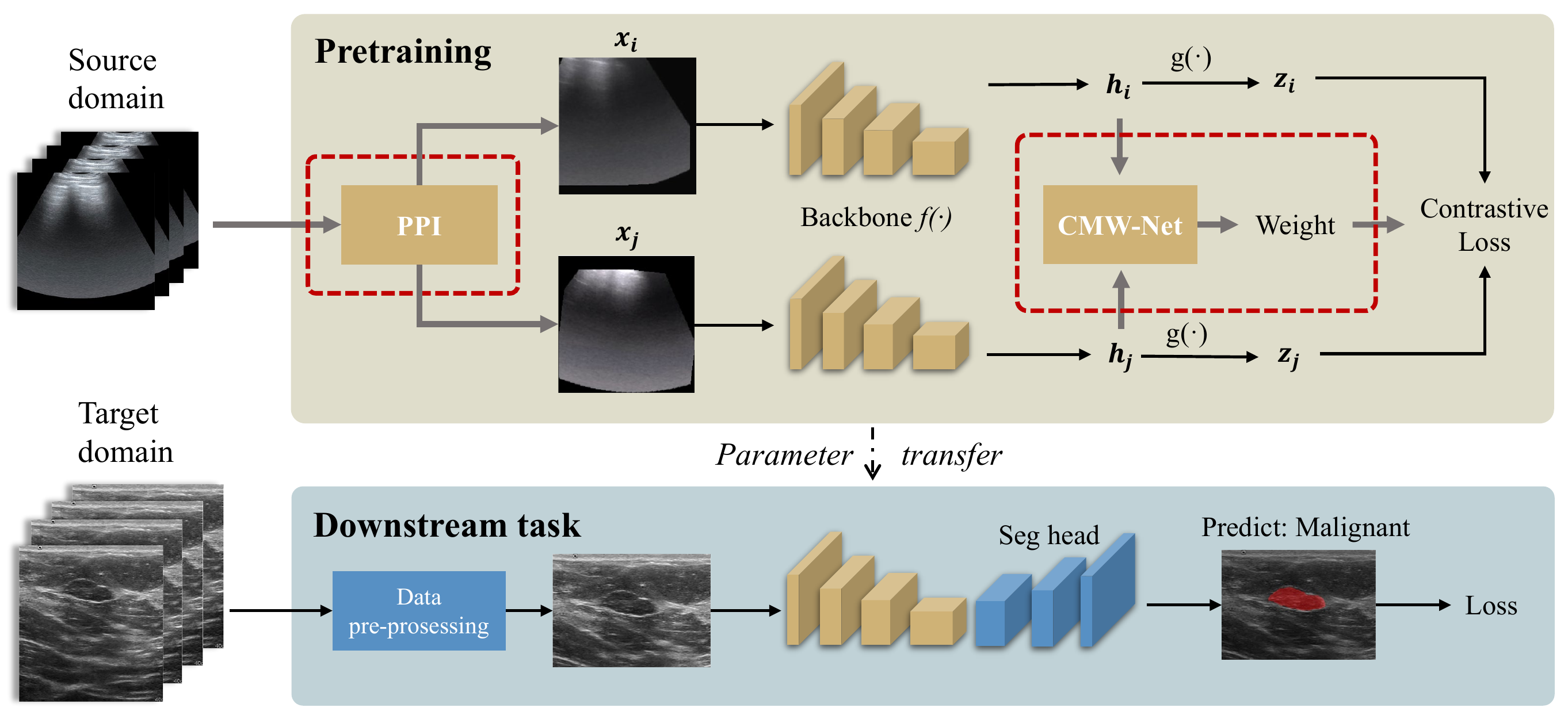}
\caption{System framework of the proposed Meta-USCL, which conducts contrastive learning with PPI module to generate positive pairs, along with the CMW-Net for weighting these sample pairs.}
\label{fig:framework}
\end{figure*}

Although PPI mitigates positive-pair dissimilarity defect and positive-pair similarity redundancy, different randomly generated positive pairs will still be different \wrt their semantic consistency and appearance difference, resulting in different degrees of contribution to learning representations. For example, the positive pairs from videos with low motion, less informative regions (\eg lung), or generated with extreme interpolation coefficients would contribute less or even do harm to learning semantically consistent mappings. This work proposes to make the training process focus more on beneficial positive pairs instead of the less helpful ones by training a meta-learning~\cite{franceschi2018bilevel,ren2018learning,shu2019meta} driven weighting network, namely \textbf{C}ontrastive \textbf{M}eta \textbf{W}eight Network (CMW-Net, Fig. \ref{fig:Motivation} (e)). The whole training framework combining PPI and CMW-Net is called Meta \textbf{U}ltra\textbf{S}ound \textbf{C}ontrastive \textbf{L}earning (Meta-USCL, see Fig. \ref{fig:framework}) in this work. The key idea of Meta-USCL is to properly generate and weight positive sample pairs with PPI and CMW-Net to make the contrastive learning benefit from learning the semantic consistency inherent in US videos.

This work is an extension of our previous USCL~\cite{chen2021uscl}, a semi-supervised contrastive learning framework that achieves significantly better performance than ImageNet pre-training. This work extends and improves the previous work in the following ways: (i) We improve the USCL to Meta-USCL, which combines meta-learning with contrastive learning to overcome the bottleneck of sample pairs with uneven quality that USCL usually encounters. (ii) We provide the interpretability of meta-weighting through visualization, and attribute the benefit brought by CMW-Net to its ability to emphasize the positive pairs that teach the model more semantic consistency of US images. (iii) More extensive experiments and analyses are presented to validate the proposed Meta-USCL on various downstream applications.

\section{Related Works}

The success of medical transfer learning is mainly attributed to the learned medical-related visual representations during the pre-training phase. In the following, we first review the works related to representation learning for medical applications, and then review the meta-weighting method, which is the key component for Meta-USCL.

\subsection{Supervised Representation Learning for Medical Applications}

For boosting the performance of DNNs on small datasets, representation learning in a supervised fashion is the earliest-developed method, among which ImageNet~\cite{deng2009imagenet} pre-training is the most popular one. The effect of ImageNet pre-training is verified by various medical tasks such as dermatologist-level classification of skin cancer \cite{esteva2017dermatologist}, recognition of Alzheimer's disease~\cite{ding2019deep}, and abnormalities with X-ray \cite{guendel2018learning}. 

Though the power of ImageNet transfer learning is robust for various medical applications~\cite{shin2016deep}, the transfer performance will be inevitably harmed by the large domain gap between natural images and medical images~\cite{tajbakhsh2016convolutional}.
For example, transferring the ImageNet representations to glioma subtype identification will be less helpful since gliomas are highly heterogeneous tumors with variable 3D imaging phenotypes~\cite{ning2021deep}, which is hardly related to any images in ImageNet. When large-scale annotated medical datasets are available, supervised pre-training directly on medical domains becomes a preferred method to bridge the domain gaps. For example, Tajbakhsh \etal \cite{tajbakhsh2015computer} designed a 2-channel image representation of emboli trained with voxel-level labeled pulmonary embolism data to better detect pulmonary embolism. Riasatian \etal \cite{riasatian2021fine} proposed to use weakly labeled whole slide images of formalin-fixed paraffin-embedded human pathology samples to learn histopathology image representations. 
Though frequently used and adopted by our previous conference paper~\cite{chen2021uscl}, traditional label supervision is abandoned in this work for its apparent drawback: the high requirement of domain-specific annotation prohibits its further scaling.

\subsection{Self-supervised Representation Learning for Medical Applications}

In the medical field, unlabeled image data are far richer than labeled ones. Pre-training medical models with artificial labels, namely self-supervised representation learning, was verified to be an effective way to train with unlabeled data~\cite{tajbakhsh2019surrogate}. Self-supervised learning algorithms~\cite{chen2019self,chen2020big,li2020self} surged recently in mainstream computer vision to learn generic visual representations since they show great potential to exploit the natural supervision inherent in the images themselves. Models Genesis \cite{zhou2019models}, a perturbed auto-encoder framework, is the milestone of the medical self-supervised learning for its comparative transfer performance to ImageNet strongly supervised counterparts. Since then, from the technologies of frame order prediction~\cite{jiao2020self}, generative learning with Restricted Boltzmann Machine~\cite{van2016combining}, Jigsaw game~\cite{navarro2021evaluating}, image context restoration~\cite{chen2019self,zhou2019models} to more advanced contrastive learning~\cite{Jianbo2020MICCAI}, self-supervised representation learning boosted the medical models' transferabilities to small datasets significantly. 

Recently, contrastive learning~\cite{Ting2020CVPR} has been a dominating self-supervised method for its outstanding performance. It aims to pull semantically consistent positive samples together in the representation space, while pushing those dissimilar negative samples away. With the fast development of contrastive learning frameworks~\cite{he2020momentum,grill2020bootstrap}, more and more works~\cite{zhou2020comparing,chen2021uscl} seek to combine properties inherent in medical data for learning better image representations. Since the US video clips are much different from individual images that contrastive learning was initially designed for, previous methods~\cite{jiao2020self,Jianbo2020MICCAI} are not effective enough to learn powerful representations from them. We regard a key factor to the success of US contrastive learning to be mining semantic consistency from video clips to generate reasonable positive pairs. Previous works~\cite{tian2020makes,xiao2020should} showed that an ideal positive pair should contain moderate appearance differences while keeping their semantics consistent. For this reason, this work considers the similarity between US frames as semantic continuity, and leverages this property to generate semantically similar samples from US videos.

\subsection{Meta-learning for Weighting Samples}

Sample weighting is an effective approach to improve the training performance when the data are diverse, \eg free-hand US videos. It increases the effect of ``good" samples' gradients on the parameter updating and decreases the effect of those not good enough. In contrastive learning, ``good" positive sample pairs denote those with higher appearance differences and lower semantic differences. Meta-weighting~\cite{ren2018learning,shu2019meta}, a meta-learning~\cite{finn2017model} technique, can achieve this target without any handcrafted weighting schemes. It is regarded as a bi-level optimization process~\cite{franceschi2018bilevel}, and aims at optimizing the weighting scheme (upper-level objective) for conducting a better learning process (lower-level objective).
To learn a better weighting scheme, there are two main meta-learning frameworks. The first one~\cite{ren2018learning} is assigning weights for each sample, and updating the weights based on the generalization error given by this weighting scheme. This method is easy-to-implement but converges slowly. The second one~\cite{shu2019meta} is weighting the samples with a weighting module, which is usually a shallow network. The parameters of the weighting module are updated to minimize the generalization error with their validation gradients.
The recent applications of meta-weighting contain the recognition of out-of-distribution samples~\cite{guo2020safe}, label noises, and training set biases~\cite{ren2018learning}. In the medical field, similar problems are still common, but only a few pieces of literature use this method to boost training efficiency~\cite{lei2021meta,wang2020meta}. Considering that some sample pairs in contrastive learning contribute more to learning desirable representations while some are not, we assign larger weights for those more helpful sample pairs. This work extends the idea of the meta-weighting module to the medical contrastive learning task, which is a novel application scenario.

\section{Methods}

\subsection{Formulation of Meta-USCL}\label{sect:formulation}

The core idea of contrastive learning methods~\cite{vu2021medaug,Jianbo2020MICCAI,arXiv2020Yuhao,sowrirajan2021moco} is instance-level discrimination, which generates sample pairs from the original dataset and trains the neural networks by recognizing similar positive pairs against dissimilar negative pairs. Meta-USCL extends the core idea of contrastive learning from two aspects: 1) generating positive sample pairs from US videos by PPI module; 2) weighting the generated positive pairs with CMW-Net to make the training more efficient. The whole framework containing Meta-USCL pre-training and parameter transferring to downstream tasks is shown in Fig. \ref{fig:framework}.

The first part of Meta-USCL is generating positive sample pairs from videos. Given a raw video $V_i$, we construct a frame set $\mathbb{F}_i^{K} \!=\!\{\textbf{f}_i^{(k)}\}^K_{k=1}$ to discard overly redundant information, where $K$ is the number of extracted images. A simple way to do this is to sample several frames per second in a video. Then, the PPI module is used to generate a positive sample pair $\mathbb{S}_i^2\!=\!\{\textbf{x}^{(1)}_i, \textbf{x}^{(2)}_i \}$ from the frame set. 

After generating a sample pair $\mathbb{S}_i^2$, a DNN backbone $f(\cdot)$ encodes them to be a representation vector pair $h_{2i-1}=f(\textbf{x}^{(1)}_i)$ and $h_{2i}=f(\textbf{x}^{(2)}_i)$. In a video batch of size $N$, there are $2N$ samples in total, where $2i-1$ and $2i$ indicate two samples generated from the video $V_i$. Subsequently, the projection head (a two-layer MLP) $g(\cdot)$ maps the representations to be feature vectors $\{\textbf{z}_{2i-1}, \textbf{z}_{2i} \}_{i=1}^N=\{g(\textbf{h}_{2i-1}), g(\textbf{h}_{2i}) \}_{i=1}^N$ specialized for contrastive learning. In this feature space, samples can better be pulled together or push apart with contrastive loss without affecting the representation quality for downstream tasks. In addition, the CMW-Net compares the representation pairs of each positive sample pair, and outputs quality scores (weights) $\{\mathcal{W}(\textbf{h}_{2i-1}, \textbf{h}_{2i})\}_{i=1}^N$ for reweighting their contrastive losses. 

The InfoNCE \cite{oord2018representation} is the most widely-used loss function for contrastive learning. It aims at minimizing the distances between positive pairs $\{\textbf{x}_{i}^{(1)}, \textbf{x}_{i}^{(2)} \}_{i=1}^N$, and maximizing the distances between negative pairs $\{\textbf{x}^{(1~or~2)}_i, \textbf{x}^{(1~or~2)}_j \},~i \! \ne \!j~,1\le i,j \le N$ in the mini-batch. This work modifies the InfoNCE to a weighted version, so that the loss for different positive sample pairs can update the model parameters differently:

\begin{align}
\mathcal{L} &= \frac{1}{2N}\sum_{i=1}^{N}\mathcal{W}(\textbf{h}_{2i-1}, \textbf{h}_{2i}) L(\textbf{z}_{2i-1}, \textbf{z}_{2i})\notag\\
&= \frac{1}{2N}\sum_{i=1}^{N}\mathcal{W}(\textbf{h}_{2i-1}, \textbf{h}_{2i})(l(2i,2i-1)+l(2i-1,2i)),
\label{eq:consistency}
\end{align}
where
\begin{equation}
l(i,j) =-\log\frac{\exp(s_{i,j}/\tau)}{\sum_{k=1}^{2N}\mathbbm{1}_{[i\ne k]}(k) \cdot exp(s_{i,k}/\tau)},
\label{eq:ll}
\end{equation}
and
\begin{equation}
s_{i,j} = \textbf{z}_i \cdot \textbf{z}_j/(\|\textbf{z}_i\| \|\textbf{z}_j\|)
\label{eq:ss}
\end{equation}
denotes the cosine similarity of a feature vector pair $\{\textbf{z}_i, \textbf{z}_j \}$. The indicator functon $\mathbbm{1}_A(x)=1$ when $x\in A$, otherwise it equals to 0. And $\tau$ is a tuning temperature parameter. 

Finally, the pre-training can be conducted with PPI-generated positive pairs and weighted InfoNCE loss. The contrastive training process learns the visual differences which would not affect the meaning of a US image. They act as the prior knowledge to assist the downstream classification or segmentation tasks.

\subsection{Positive Pair Interpolation Module}\label{sect:SPG}

The typical contrastive learning paradigm~\cite{Ting2020CVPR} generates positive pairs based on random augmentation and cannot perform well for US videos, because they were primarily designed for image datasets and will meet the three problems introduced in Sec. \ref{sec:introduction}. 

\begin{figure}[t]
\centering
\includegraphics[width=0.99\linewidth]{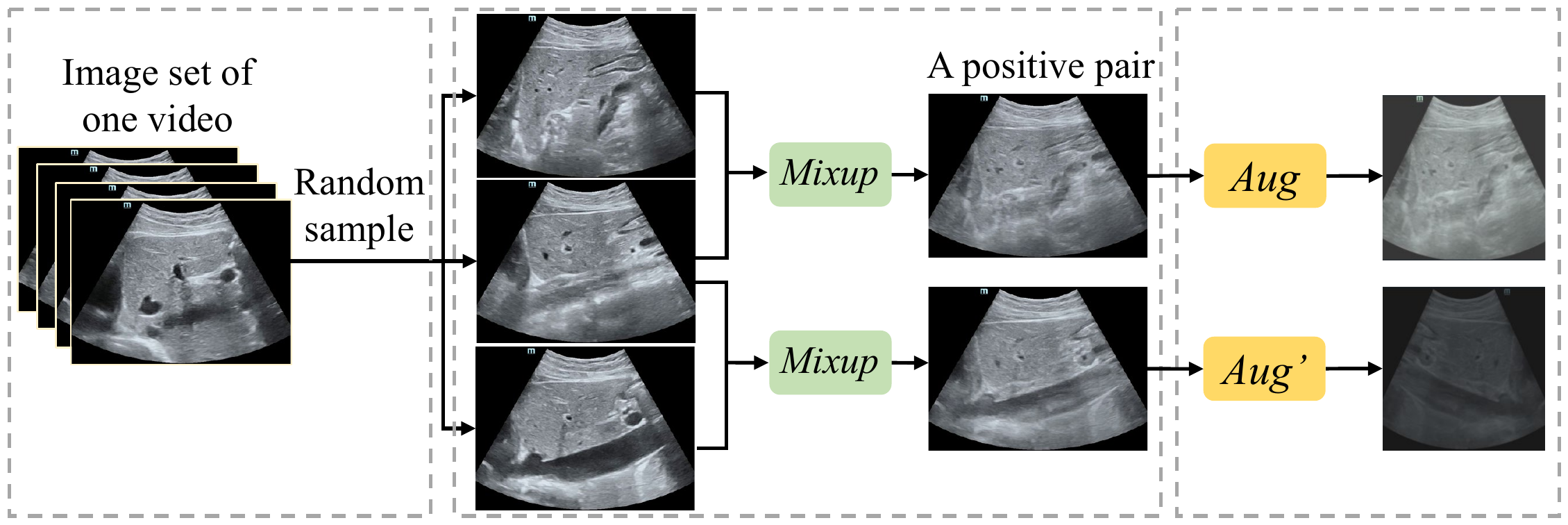}
\caption{Illustration of the PPI module. It contains three steps: 1) randomly sampling three frames from a video; 2) taking the middle frame as the ``anchor" and using mix-up operation to generate two new samples; 3) augmenting the samples randomly to get the positive sample pair.} 
\label{fig:positivesampleselection}
\end{figure}

First, to prevent the negative-pair similarity conflict problem, the PPI module generates only one positive sample pair from one video. Secondly, it uses a mix-up~\cite{mixup2018ICLR} operation to generate semantically consistent positive pairs, which mitigates the positive-pair dissimilarity defect and similarity redundancy problems. The process of PPI is visualized in Fig. \ref{fig:positivesampleselection}. In the first step, it randomly selects three frames $\widehat{\textbf{x}}_i^{(1)}, \widehat{\textbf{x}}_i^{(2)}, \widehat{\textbf{x}}_i^{(3)}$ in chronological order from the frame set $\mathbb{F}_i^{K}$ of video $i$. And then it regards the middle frame $\widehat{\textbf{x}}_i^{(2)}$ in temporal as an ``anchor" frame, interpolates it with the other two perturbation frames $\widehat{\textbf{x}}_i^{(1)}, \widehat{\textbf{x}}_i^{(3)}$ respectively:

\begin{equation}
\left\{\begin{array}{l}
\textbf{x}_i^{(1)} = \xi_1 \widehat{\textbf{x}}_i^{(2)} + (1-\xi_1) \widehat{\textbf{x}}_i^{(1)}\\
\textbf{x}_i^{(2)} = \xi_2 \widehat{\textbf{x}}_i^{(2)} + (1-\xi_2) \widehat{\textbf{x}}_i^{(3)}
\end{array},\right.
\label{eq:mixup}
\end{equation}
where $\xi_1, \xi_2 \sim Beta(\alpha, \beta)$, and $\alpha, \beta$ are parameters of $Beta$ distribution. On the one hand, positive pairs are random offsets from the anchor image to the perturbation images, ensuring that they share the common semantic information from the anchor image, which overcomes the positive-pair dissimilarity defect problem. On the other hand, the sampling interval for a frame set $\mathbb{F}_i^{K}$ results in low probability for PPI to sample three similar frames $\{\widehat{\textbf{x}}_i^{(k)}\}_{k=1}^3$ which are too close in appearance, which prevents the positive-pair similarity redundancy problem. The generated two positive samples $\textbf{x}_i^{(1)}, \textbf{x}_i^{(2)}$ would be further randomly augmented to gain more abundant visual differences. 

The PPI module generates positive pairs and regards samples from different videos as negative pairs. The benefits of the PPI can be summarized as follows: 1) Interpolation makes every point in the feature convex hull enclosed by the cluster boundary possible to be sampled, making the cluster cohesive as a whole; 2) Positive pairs generated with Eq. (\ref{eq:mixup}) have guaranteed semantic consistency and appearance discrepancies to make the training more stable and efficient.

\begin{table*}[]
\centering
\caption{Statistics of the {US-4} dataset containing 4 video-based sub-datasets. The total number of images is 39,021, uniformly sampled from 1366 videos. Most videos contain 10$\sim$50 similar images, which ensures the good property of semantic clusters.}
\label{tab:US-4}
\setlength{\tabcolsep}{5mm}
\renewcommand{\arraystretch}{1.2}
\begin{tabular}{ccccccc}
	\toprule
	Sub-dataset & Organ &  Image size & Depth & Frame rate & Videos & Images \\ 
	\midrule
	Butterfly~\cite{ButterflyData} & Lung & 658$\times$738 & - & 23Hz & 22 & 1533   \\
	CLUST~\cite{Somphone2014MICCAI} & Liver & 434$\times$530 & - &  19Hz & 63   & 3150    \\ 
	{Liver Fibrosis} & Liver & 600$\times$807 & $\sim$8cm & 28Hz & 611 & 27504   \\ 
	{COVID19-LUSMS} & Lung & 747$\times$747 & $\sim$10cm & 17Hz & 670 & 6834   \\ 
	\bottomrule
\end{tabular}
\end{table*}

\subsection{Learning to Weight Sample Pairs}\label{sect:metaLearning}

For US videos that are too long, PPI may still generate dissimilar positive pairs due to the random mix-up coefficients $\xi_1, \xi_2$ of the $Beta$ distribution. In addition, videos collected by different sonographers from different organs have different sequential characteristics, making them contribute to the representation learning differently in quality. So it is challenging for PPI to solve the three problems thoroughly. Sample weighting with CMW-Net is a promising method to make the model learn more with desirable sample pairs and further mitigate these problems. 

The CMW-Net is mainly inspired by~\cite{shu2019meta}, which proposed a sample weighting net to discover the important samples for training. This work extends this method and proposes to obtain weights for each positive pair in contrastive learning. Let $\Theta_{m}, \Theta_{c}$ denote the parameters of the main pre-training network and the CMW-Net, respectively. This weighting network takes the representation vectors $\textbf{h}_{2i-1}, \textbf{h}_{2i}$ of a positive pair as the input, and outputs corresponding quality score $\mathcal{W}(\textbf{h}_{2i-1}, \textbf{h}_{2i}, \Theta_{c})$. The weighted training loss $\mathcal{L}^{train}(\Theta_{m},\Theta_{c})$ can be defined as Eq. \ref{eq:consistency} on the training set.
The aim of representation learning is to minimize the weighted training loss \wrt the main network parameter $\Theta_{m}$ given CMW-Net parameter $\Theta_{c}$.

\begin{equation}
\Theta^*_{m}(\Theta_{c}) = \mathop{\arg\min}_{\Theta_{m}} \mathcal{L}^{train}(x,\Theta_{m},\Theta_{c}).
\label{eq:lower}
\end{equation}

Note that $\Theta^*_{m}$ can be regarded as a function of $\Theta_{c}$ since it acts as part of the learning scheme. The parameters $\Theta_{c}$ decides how the weight of each sample pair is calculated. We use the validation loss as a guide for the CMW-Net to find parameters with better generalization. The updating of $\Theta_{c}$ is based on the unweighted loss $\mathcal{L}^{valid}(\Theta^*_{m}(\Theta_{c}))$ calculated on the validation sample pairs with optimal main network parameter $\Theta^*_{m}$:

\begin{align}
\Theta^*_{c} &= \mathop{\arg\min}_{\Theta_{c}} \mathcal{L}^{valid}(\Theta^*_{m}(\Theta_{c}))\notag\\
&= \mathop{\arg\min}_{\Theta_{c}} \frac{1}{2N}\sum_{i=1}^{N}L^{valid}(\textbf{z}_{i}^{(1)}, \textbf{z}_{i}^{(2)}, \Theta^*_{m}(\Theta_{c})).
\label{eq:upper}
\end{align}

The problem for deriving optimal parameters of the two networks can be understood as a nested optimization, where the lower-level objective is the minimization of training loss \wrt $\Theta_{m}$ (Eq. \ref{eq:lower}) while the upper-level objective is the minimization of validation loss \wrt $\Theta_{c}$ (Eq. \ref{eq:upper}).\\

\textbf{Parameter Updating Procedure.} To solve the above optimization problem, we design an online updating algorithm inspired by \cite{shu2019meta}. The core idea is to iteratively update the parameters of the two targets with the following three steps: (1) getting feedback about the current weights; (2) updating the weighting strategy; and (3) updating the network parameters. In the iteration $t$, the algorithm first updates the main model parameters
\begin{align}
\hat{\Theta}^{(t)}_{m}(\Theta^{(t)}_{c}) =
    \Theta_{m}^{(t)} - \alpha_1 \nabla_{\Theta_{m}} \mathcal{L}^{train}(\Theta_{m},\Theta_{c}^{(t)}) \left.\right|_{\Theta_{m}^{(t)}},
\end{align}
with one step of gradient descent, where $\alpha_1$ is the learning rate of the main model. This step is performed in a mini-batch of training data. Its function is to get feedback for the weights given by CMW-Net. 

After taking the derivative of the meta (validation) loss \wrt the $\Theta_{c}$, we can take the second step to move the $\Theta_{c}^{(t)}$ along the meta objective gradient on the validation data
\begin{align}
\Theta_{c}^{(t+1)} =
    \Theta_{c}^{(t)} - \alpha_2 \nabla_{\Theta_{c}} \mathcal{L}^{valid}(\hat{\Theta}_{m}^{(t)}(\Theta_{c})) \left.\right|_{\Theta_{c}^{(t)}},
\end{align}
with $\alpha_2$ as the learning rate of the CMW-Net. 

Finally, the updated $\Theta_{c}^{(t+1)}$ is used to generate the weights of this batch of sample pairs, and obtain the updated main model parameters with the third step of gradient descent:

\begin{align}
\Theta_{m}^{(t+1)}(\Theta_{c}^{(t+1)}) =
    \Theta_{m}^{(t)} - \alpha_1 \nabla_{\Theta_{m}} \mathcal{L}^{train}(\Theta_{m},\Theta_{c}^{(t+1)}) \left.\right|_{\Theta_{m}^{(t)}}.
\end{align}

All these three steps are conducted in an iterative fashion. They update the parameters of CMW-Net and main model alternatively.

\section{Experimental Setup}\label{sect:expSetup}

This section introduces the US-4 dataset for conducting meta-USCL, along with the implementation details of the pre-training and fine-tuning processes.

\begin{table*}[]
\centering
\scriptsize
\caption{Evaluation on POCUS dataset for two main components of Meta-USCL: PPI and meta-reweighting. ``Params" and ``MACs" denotes the network parameters and forward computations. ``$\times0.5$, $\times1.0$, $\times1.5$, $\times2.0$" means the width of ShuffleNet-v2. All results are 5-fold cross validation accuracy (\%).}
\renewcommand{\arraystretch}{1.2} 
\resizebox{\textwidth}{40mm}{ 
\begin{tabular}{cccccccccccc}
\toprule
\multirow{2}{*}{Model} & \multirow{2}{*}{Params} & \multirow{2}{*}{MACs} & \multirow{2}{*}{Image size} & \multirow{2}{*}{From Scratch} & \multirow{2}{*}{ImageNet} &  \multicolumn{6}{c}{Meta-USCL} \\
& & & & & & SimCLR & meta+SimCLR & $S_1$  & $S_2$  & $S_3$ (USCL) & meta+$S_3$ \\
\midrule
\multirow{3}{*}{ShuffleNet-v2 $\times0.5$} & \multirow{3}{*}{1.4M} & \multirow{3}{*}{4.5M} & $64^2$ & $68.9 \pm 1.1$ & $77.2 \pm 0.7$ & $78.6 \pm 1.1$ & $83.3 \pm 0.5$ & $81.2 \pm 0.7$ & $84.8 \pm 0.7$ & $86.4 \pm 1.0$ & $87.8 \pm 1.2$ \\
&  &  & $128^2$ & $71.1 \pm 1.7$ & $78.8 \pm 1.3$ & $80.0 \pm 1.6$ & $84.5 \pm 0.7$ & $83.3 \pm 0.5$ &  $86.4 \pm 0.6$ & $87.9 \pm 0.9$ & $88.3 \pm 1.4$ \\
&  &  & $224^2$ & $74.6 \pm 2.2$ & $79.4 \pm 1.2$ & $82.1 \pm 1.5$ & $85.2 \pm 0.4$ & $84.7 \pm 1.3$ & $87.0 \pm 0.4$ & $88.8 \pm 0.9$ & $90.1 \pm 0.6$ \\
\midrule
\multirow{3}{*}{ShuffleNet-v2 $\times1.0$} & \multirow{3}{*}{2.3M} & \multirow{3}{*}{13.2M} & $64^2$ & $72.4 \pm 0.9$ &  $78.5 \pm 1.9$ & $80.3 \pm 1.2$ & $82.2 \pm 0.2$ & $81.5 \pm 1.2$ & $83.9 \pm 1.3$ & $86.1 \pm 1.4$ & $87.3 \pm 0.5$ \\
&  &  & $128^2$ & $74.7 \pm 1.9$ & $79.9 \pm 1.7$ & $81.3 \pm 1.0$ & $85.0 \pm 0.8$ & $82.8 \pm 1.5$ & $85.2 \pm 0.7$ & $85.9 \pm 1.1$ & $88.9 \pm 0.4$ \\
&  &  & $224^2$  & $75.7 \pm 1.3$ &  $80.6 \pm 1.6$ & $81.3 \pm 1.6$ & $85.7 \pm 0.6$ & $83.2 \pm 1.4$ & $85.6 \pm 1.7$ & $89.2 \pm 1.8$ & $91.3 \pm 0.6$ \\
\midrule
\multirow{3}{*}{ShuffleNet-v2 $\times1.5$} & \multirow{3}{*}{3.5M} & \multirow{3}{*}{25.7M}  & $64^2$ & $74.1 \pm 0.6$ &   $78.6 \pm 0.7$ & $78.8 \pm 0.9$ & $82.9 \pm 0.9$ & $81.0 \pm 0.8$   & $84.3 \pm 0.9$ & $86.6 \pm 0.7$ & $88.5 \pm 0.9$ \\
&  &  & $128^2$ & $76.6 \pm 1.7$ & $79.3 \pm 1.6$ & $80.4 \pm 0.7$ & $83.5 \pm 0.6$ & $82.1 \pm 1.3$ & $86.2 \pm 1.0$ & $88.4 \pm 0.9$ & $89.6 \pm 0.5$ \\
&  &  & $224^2$ & $79.7 \pm 1.3$ & $81.5 \pm 0.8$ & $81.9 \pm 0.4$ & $85.5 \pm 0.8$ & $83.4 \pm 1.1$ & $86.9 \pm 0.8$ & $89.3 \pm 0.6$ & $91.2 \pm 0.3$ \\
\midrule
\multirow{3}{*}{ShuffleNet-v2 $\times2.0$} & \multirow{3}{*}{7.4M} & \multirow{3}{*}{50.4M}  & $64^2$ & $75.1 \pm 0.8$ &  $79.8 \pm 0.8$  & $78.7 \pm 0.6$ & $82.8 \pm 0.6$ & $80.0 \pm 0.5$   & $84.1 \pm 1.1$ & $86.1 \pm 0.9$ & $86.5 \pm 0.9$ \\
&  &  & $128^2$ & $77.3 \pm 1.2$ & $79.0 \pm 1.0$ & $80.7 \pm 1.3$ & $83.4 \pm 1.3$ & $81.8 \pm 0.8$ & $86.1 \pm 0.8$ & $88.0 \pm 0.3$ & $89.5 \pm 0.7$ \\
&  &  & $224^2$ & $80.1 \pm 1.9$ & $81.2 \pm 0.7$ & $80.5 \pm 0.9$ & $84.7 \pm 0.6$ & $84.4 \pm 1.0$ & $86.6 \pm 1.3$ & $88.9 \pm 0.5$ & $90.3 \pm 0.6$ \\
\midrule
\multirow{3}{*}{ResNet18} & \multirow{3}{*}{11.7M} & \multirow{3}{*}{149.2M}  & $64^2$ & $ 84.1 \pm 0.8$ & $81.9 \pm 0.8$ & $82.4 \pm 0.8$ & $84.1 \pm 0.9$ & $84.1 \pm 0.6$ & $87.5 \pm 1.3$ & $90.0 \pm 1.2$ & $90.9 \pm 0.7$ \\
&  &  & $128^2$ & $83.6 \pm 1.4$ & $83.7 \pm 0.6$ & $85.9 \pm 1.5$ & $88.3 \pm 0.9$ & $88.0 \pm 0.7$ & $89.5 \pm 1.1$ & $92.1 \pm 0.9$ & $93.1 \pm 0.6$ \\
&  &  & $224^2$ & $84.3 \pm 0.9$ &  $84.2 \pm 0.9$ & $87.5 \pm 1.1$ & $91.0 \pm 0.6$ & $90.8 \pm 0.5$ & $92.3 \pm 0.8$ & $93.2 \pm 0.3$ & $94.6 \pm 0.4$ \\
\midrule
\multirow{2}{*}{ResNet34} & \multirow{2}{*}{21.8M} & \multirow{2}{*}{300.5M} & $64^2$ & $83.9 \pm 0.7$ & $78.9 \pm 0.7$ & $78.9 \pm 0.7$ & $80.2 \pm 1.2$ & $81.5 \pm 1.0$ & $84.7 \pm 1.3$ & $87.7 \pm 0.5$ & $88.3 \pm 1.1$ \\
& & & $128^2$ & $84.6 \pm 0.9$ & $83.4 \pm 0.5$ & $83.2 \pm 1.5$ & $86.9 \pm 0.8$ & $86.5 \pm 0.9$ & $87.7 \pm 0.7$ & $90.4 \pm 0.9$ & $91.5 \pm 0.8$ \\
\midrule
\multirow{2}{*}{ResNet50} & \multirow{2}{*}{25.6M} & \multirow{2}{*}{338.4M} & $64^2$ & $82.9 \pm 1.0$ & $78.7 \pm 0.3$ & $84.2 \pm 0.7$ & $80.3 \pm 0.7$ & $85.5 \pm 1.4$ & $85.5 \pm 1.3$ & $88.7 \pm 0.6$ & $89.0 \pm 1.4$ \\
& & & $128^2$ & $84.4 \pm 0.6$ & $84.7 \pm 0.8$ & $86.3 \pm 0.5$ & $87.6 \pm 0.8$ & $86.8 \pm 1.2$ & $87.5 \pm 0.7$ & $91.2 \pm 0.3$ & $92.3 \pm 1.0$ \\
\bottomrule
\end{tabular}}
\label{tab:ablation}
\end{table*}

\subsection{Pre-training Dataset: US-4}

In this work, we construct a US pre-training dataset named US-4. It is collected from four convex probes~\cite{born2020pocovid} US video datasets involving two scan regions (\ie lung and liver). The attributes (\eg depth and frame rate) of the US-4 dataset are described in Tab. \ref{tab:US-4}. Among the four sub-datasets, Liver Fibrosis and COVID19-LUSMS datasets are collected by our local sonographers~\cite{li2020self,liu2020semi} with Resona 7T ultrasound system. The frequency is FH 5.0, and the pixel size is 0.101mm - 0.127mm. Butterfly~\cite{ButterflyData} and CLUST~\cite{Somphone2014MICCAI} are two public sub-datasets. In order to generate a diverse and sufficiently large dataset, images are selected from original videos with a suitable sampling interval. We extract three samples per second for each video, ensuring the sampling interval $I>5$ so that US-4 contains sufficient but not redundant information about videos. This results in 1366 videos and 39,021 images. US-4 is relatively balanced regarding the number of images in each video, where most videos contain tens of US images.

\subsection{Training Setup}

Meta-USCL proposed in this work is implemented with PyTorch 1.8.0. The model training can be split into two parts: pre-training and fine-tuning. In the pre-training phase, 80\% of data are used for training, and 20\% are used for validation. The batch size of 32 is adopted to conduct the training. We conduct contrastive learning with $224\times224$ images for the POCUS pneumonia detection and the UDIAT-B breast tumor segmentation tasks. For BUI breast cancer diagnosis, we use $64\times64$ images. The key reason for using a small image size for transferring to this classification task is that the BUI dataset mainly contains small images, which will be elaborated on in Sec. \ref{sect: comparison}. For the experiments without meta-learning, the optimizer is Adam, with a learning rate $\alpha_1=10^{-3}$ and an $L2$ regularization (weight decay) strength of $10^{-4}$. The temperature parameter $\tau$ equals 0.5 as in \cite{Ting2020CVPR}. For the experiments with meta-learning, the optimizer for the main model is replaced with SGD with a learning rate of $\alpha_1=0.1$. The learning rate for CMW-Net is $\alpha_2=6\times 10^{-5}$ as in the \cite{shu2019meta}. In each case, the images are augmented with random cropping followed by transformation to the initial size, random flipping, and random color jittering to make the model robust to different image appearances from different imaging systems. As important hyper-parameters, the batch size and the strengths for data augmentations are discussed in detail (see Sec. \ref{sect:analysis}).

As for fine-tuning, we consider downstream classification and segmentation tasks. For classification tasks, we use a consistent batch size of 128 with the same image sizes as pre-training. The optimization method is set to be Adam with a learning rate of 0.01, training 30 epochs. Images are augmented with randomly cropping, resizing, and flipping. Please note that only the last three layers of pre-trained models are tuned in the fine-tuning phase of classification. For the segmentation task, we discard the final classification layer and use the pre-trained DNN backbone to construct a Mask-RCNN~\cite{kaiming2020TPAMI} model, and fine-tune it on the target dataset. The segmentation is trained with the batch size of 2 in the original image size. SGD method with learning rate $5\times10^{-3}$, momentum 0.9, $L2$ regularization strength $10^{-4}$ is used for optimization. Only random flipping is used for data augmentation. All fine-tuning experiments are repeated 5 times. The mean and standard deviation of the results are reported to ensure the reliability of the results.

\section{Experimental Results and Analysis}

In this section, we start with validating each technique proposed in this work, including PPI and meta-weighting. And then, we continue with the comparative experiments for comparing the transferability of Meta-USCL pre-trained models to other representation learning methods. Finally, we explore some properties of Meta-USCL based on the US-4 dataset and show the potential to improve its performance further.

\begin{table*}[]
\centering
\scriptsize
\caption{Comparative fine-tuning results (\%) on POCUS classification dataset. F1 for three-class classification is defined as the average F1 of all classes. Significance is reported between our methods and ImageNet pre-training ($>$***: $p<0.0001$).}
\renewcommand{\arraystretch}{1.2}
\begin{tabular}{c c c c c c c c}
\toprule
\multirow{2}{*}{Method} & \multicolumn{2}{c}{COVID-19} & \multicolumn{2}{c}{Pneumonia}  & \multirow{2}{*}{normal/abnormal} & \multirow{2}{*}{F1} & \multirow{2}{*}{Accuracy} \\ \cline{2-5}
    & \multicolumn{1}{c}{Sensitivity} & Specificity & \multicolumn{1}{c}{Sensitivity} & Specificity & & & \\ \midrule
From Scratch & \multicolumn{1}{l}{$80.8\pm1.3$} & $89.5\pm1.1$ & \multicolumn{1}{l}{$83.6\pm1.5$} & $94.4\pm1.4$ & $86.7\pm1.6$ & $83.9\pm0.6$ & $84.3\pm0.9$ \\
ImageNet & \multicolumn{1}{l}{$83.2\pm0.9$} & $89.9\pm1.2$ & \multicolumn{1}{l}{$86.6\pm3.3$} & $93.4\pm1.1$ & $85.3\pm0.7$ & $83.2\pm0.5$ & $84.2\pm0.9$ \\
US-4 supervised & \multicolumn{1}{l}{$82.1\pm 2.1$} & $91.3\pm 1.8$ & \multicolumn{1}{l}{$86.1\pm3.1$} & $91.8\pm1.5$ & $86.6\pm0.8$ & $84.7\pm0.7$ & $85.0\pm0.6$ \\
SimCLR & \multicolumn{1}{l}{$81.2\pm1.3$} & $89.6\pm0.9$ & \multicolumn{1}{l}{$88.4\pm2.2$} & $98.2\pm0.3$ & $86.0\pm0.5$ & $86.2\pm1.0$ & $86.3\pm0.8$ \\
MoCo v2 & \multicolumn{1}{l}{$79.0\pm2.0$} & $90.2\pm0.7$ & \multicolumn{1}{l}{$83.5\pm2.6$} & $97.7\pm0.6$ & $86.5\pm0.6$ & $84.1\pm0.7$ & $84.8\pm0.3$ \\
Jiao et. al. & \multicolumn{1}{l}{$83.8\pm1.7$} & $87.7\pm1.1$  & \multicolumn{1}{l}{$78.5\pm2.4$} & $95.7\pm0.4$ & $87.1\pm0.7$  & $83.9\pm0.8$ & $84.6\pm0.5$ \\
\hline
Meta+SimCLR & $81.1\pm1.4$ & $92.7\pm1.0$ & $89.0\pm0.7$ & $98.4\pm0.1$ & $93.7\pm0.4$ & $90.6\pm0.3$ & $91.0\pm0.6$ ($>$***) \\
USCL & $90.2\pm0.5$ & $92.9\pm1.4$ & $90.8\pm1.3$ & $\bm{98.8}\pm0.2$ & $92.4\pm0.6$ & $93.0\pm0.5$ & $93.2\pm0.3$ ($>$***) \\
Meta-USCL & \multicolumn{1}{l}{$\bm{91.5}\pm1.4$} & $\bm{94.7}\pm0.6$ & \multicolumn{1}{l}{$\bm{92.9}\pm1.2$}  & $\bm{98.8}\pm0.4$ & $\bm{95.2}\pm0.3$ & $\bm{94.1}\pm0.6$ & $\bm{94.6}\pm0.4$ ($>$***) \\ \bottomrule
\end{tabular}
\label{tab:pocus_compare}
\end{table*}

\subsection{Validation of Key Components}\label{sect:ablations}

We first report the experimental results validating different components of Meta-USCL (Tab. \ref{tab:ablation}). All pre-trained models are fine-tuned on the POCUS~\cite{born2020pocovid} pneumonia detection dataset. The POCUS is a widely used lung convex-probe US dataset for COVID-19 detection, consisting of 2116 frames from 140 videos for three classes (\ie 655 frames from 58 videos for COVID-19, 349 frames from 24 videos for bacterial pneumonia, and 1112 frames from 58 frames for healthy controls). The video data were originally collected by Born \etal~\cite{born2020pocovid} from online websites and pre-processed for the subsequent algorithm development. We uniformly split POCUS into five folds \wrt videos in different classes and conduct a 5-fold cross-validation to ensure that the frames of a single video are present within a single fold only.

In this experiment, we test different DNN architectures (\ie ResNet~\cite{he2016deep}, ShuffleNet-v2~\cite{ma2018shufflenet}) with different layers. First, ResNet is one of the most common network architectures for medical image analysis. Second, ShuffleNet-v2 is a practical choice for deploying DNN models to clinical US systems for its low memory footprint and high inference speed. All the architectures are chosen with increasing sizes (\ie from 1M to 25M parameters). We also test 5 variants of Meta-USCL, namely SimCLR, $S_1$, $S_2$, $S_3$, and meta+$S_3$. ``SimCLR" represents pure SimCLR trained with US-4 as an image dataset. Further, the $S_1$ variant only chooses one frame from a video and augments it twice to be a positive pair to avoid negative-pair similarity conflict, while $S_2$ chooses two frames from a video and augments them to gain richer visual features. Finally, $S_3$ (the USCL proposed by our preceding work~\cite{chen2021uscl}), is the method of using the PPI module to generate sample pairs with three frames in a video. Meta+$S_3$ adds extra CMW-Net to $S_3$ for reweighting. As shown in Tab. \ref{tab:ablation}, there are a few noteworthy results: 1) Training from scratch gives undesirable classification accuracy. 2) Transfer from ImageNet demonstrates consistent and significant improvements for ShuffleNet v2, but negative transfer appears for ResNet. This result is consistent with~\cite{born2020pocovid}, indicating that not all pre-trained models are beneficial. 3) With more and more sophisticated design for contrastive learning (from SimCLR to $S_3$), representation learned with the US-4 dataset becomes more and more powerful, finally outperforms ImageNet pre-training significantly. 4) Meta-weighting improves both SimCLR and USCL, implicating that positive pairs in contrastive learning usually do not contribute to the model training equally.

This experiment demonstrates that the pre-training on ImageNet does not bring consistent and stable improvement for the downstream performance. However, it is also hard for simple medical adaptation of self-supervised methods like SimCLR to learn satisfying US medical visual representations because the US videos have different properties from individual images, and US-4 is far smaller than regular pre-training datasets of traditional natural images. Therefore, more targeted designs for the US are needed. From SimCLR to $S_1$, the negative-pair similarity conflict problem is tackled. $S_2$ makes use of the semantic similarity between the video frames. It regards the appearance differences between frames as a perturbation, acting as a kind of natural data augmentation to promote the training process. $S_3$ uses the same idea as $S_2$ but further mitigates positive-pair similarity redundancy and positive-pair dissimilarity defect problems with PPI. The pre-training performances improve monotonically by solving negative-pair similarity conflict ($S_1$), enriching features ($S_2$), and solving positive-pair similarity redundancy and positive-pair dissimilarity defect ($S_3$) incrementally. In addition, the meta-learning module learns how important the positive sample pairs are by CMW-Net, and assigns different weights for different positive pairs in a batch. The experiment shows that the $S_3$ setting is still not perfect, and CMW-Net still improves $S_3$ by 0.3\% to 3.0\% with different DNN architectures. The reason why meta-weighting works will be elaborated in Sec. VI.

\subsection{Comparative Experiments on Various Downstream Tasks}\label{sect: comparison}

In comparative experiments, we select several supervised (ImageNet pre-training, supervised pre-training using the categorical labels of US-4) and self-supervised (SimCLR~\cite{Ting2020CVPR}, MoCo v2~\cite{chen2020improved}, Jiao \etal~\cite{jiao2020self} ) methods to compare the transfer performance on CAD and segmentation tasks. All experiments are conducted with ResNet18. We use t-tests to evaluate the significance of performance improvement between our methods to ImageNet.

\begin{table*}[]
\centering
\caption{Comparative fine-tuning results (\%) on BUI classification dataset. Significance is reported between our methods and ImageNet pre-training (***: $p<0.001$; $>$***: $p<0.0001$).}
\setlength{\tabcolsep}{4.0mm}
\renewcommand{\arraystretch}{1.2}
\begin{tabular}{cccccc}
\toprule
\multirow{2}{*}{Method} & \multirow{2}{*}{Precision} & \multirow{2}{*}{Recall} & \multirow{2}{*}{F1} & \multirow{2}{*}{AUC} & \multirow{2}{*}{Accuracy}  \\
& & & & & \\ \midrule
From Scratch & $81.6\pm3.8$ & $80.0\pm2.8$ & $80.8\pm2.6$ & $69.3\pm4.1$ & $77.2\pm3.4$ \\
ImageNet & $75.0\pm3.5$ & $87.3\pm2.7$ & $80.7\pm2.2$  & $76.6\pm3.1$ & $78.3\pm2.2$ \\
US-4 supervised & $80.6\pm3.2$ & $86.3\pm4.2$ & $83.2\pm2.8$ & $77.9\pm4.5$ & $79.2\pm3.3$ \\
SimCLR & $87.9\pm2.1$ & $82.7\pm1.7$ & $85.2\pm2.1$ & $85.8\pm3.0$ & $82.8\pm2.8$ \\
MoCo v2 & $\bm{93.7}\pm2.1$ & $78.7\pm2.0$ & $85.5\pm1.0$ & $86.2\pm2.4$ & $84.0\pm3.2$ \\
Jiao et. al. & $75.8\pm3.7$ & $\bm{94.0}\pm1.6$ & $83.9\pm2.6$ & $74.1\pm4.6$ & $78.4\pm2.7$ \\
\hline
Meta+SimCLR & $88.9\pm1.2$ & $88.3\pm2.4$ & $88.6\pm2.3$ & $90.1\pm2.1$ ($>$***) & $87.0\pm1.9$ ($>$***) \\
USCL & $90.3\pm1.0$ & $87.3\pm2.4$ & $88.8\pm2.2$ & $\bm{90.8}\pm1.8$ ($>$***) & $86.8\pm1.7$ ($>$***) \\
Meta-USCL & $89.4\pm2.1$ & $89.6\pm0.9$ & $\bm{89.5}\pm1.3$ & $89.2\pm3.2$ (***) & $\bm{87.4}\pm1.7$ ($>$***) \\ \bottomrule
\end{tabular}
\label{tab:bui_compare}
\end{table*}

\begin{table}[]
\centering
\caption{Impact of image size on the BUI dataset.}
\renewcommand{\arraystretch}{1.0}
\small
\begin{tabular}{ccccc}
\toprule
Image size & $64\times64$ & $128\times128$ & $224\times224$ \\
\midrule
Acc (\%) & $89.2\pm3.2$ & $88.9\pm3.5$ & $88.4\pm3.6$  \\
\bottomrule
\end{tabular}
\label{tab:img_size_bui}
\end{table}

\textbf{POCUS Pneumonia Detection.} We show the sensitivity and specificity of the two abnormalities in POCUS. It can be observed from Tab. \ref{tab:pocus_compare} that the Meta-USCL achieves the highest sensitivities for both COVID-19 and Pneumonia. It is also the only method to reach the normal/abnormal and overall accuracy above 90\%. In addition, some interesting results can be noticed. Firstly, training DNNs from scratch only gets 84.3\% total accuracy and low sensitivity for COVID-19, but some pre-trained models even perform worse (\eg ImageNet). This result is consistent with the previous literature~\cite{born2020pocovid}, reporting that pre-training is not necessarily helpful in some cases. For ImageNet, its domain has a large gap to the POCUS dataset, which may lead to negative transfer. 
With Meta-USCL, we can achieve performance up to about 95\%, which is higher than the 89\% reported in \cite{born2020pocovid} and 94\% in \cite{chen2021uscl}. The result shows that our method appears to be a new SOTA.

\begin{table}[]
\centering
\caption{Comparative fine-tuning results (\%) on UDIAT-B segmentation dataset. Significance is reported between our methods and ImageNet pre-training (n.s.: No Significance; *: $p<0.05$).}
\renewcommand{\arraystretch}{1.2}
\scriptsize
\begin{tabular}{cccc}
\toprule
\multirow{2}{*}{Method} & \multirow{2}{*}{PPV} & \multirow{2}{*}{Sensitivity} & \multirow{2}{*}{Dice} \\ & & & \\ \midrule
From Scratch & $81.5\pm2.7$ & $74.7\pm3.7$ & $77.9\pm3.8$ \\
ImageNet & $82.0\pm3.2$ & $80.8\pm2.8$ & $81.4\pm3.4$ \\
US-4 supervised & $77.2\pm3.4$ & $79.3\pm3.1$ & $78.2\pm3.1$ \\
SimCLR & $77.4\pm3.7$ & $80.8\pm3.9$ & $79.1\pm2.5$ \\
MoCo v2 & $72.6\pm4.3$ & $78.6\pm1.9$ & $75.4\pm3.1$ \\
Jiao et. al. & $63.5\pm4.7$ & $73.1\pm3.8$ & $67.9\pm5.8$ \\
\hline
Meta+SimCLR & $\bm{92.1}\pm2.4$ & $73.3\pm2.0$ & $83.6\pm2.7$ (n.s.) \\
USCL & $88.2\pm2.1$ & $80.5\pm1.2$ & $84.1\pm1.9$ (*) \\
Meta-USCL & $87.0\pm3.2$ & $\bm{82.2}\pm2.1$ & $\bm{84.5}\pm2.9$ (*) \\ \bottomrule
\end{tabular}
\label{tab:udiat_compare}
\end{table}

\textbf{BUI Breast Cancer Classification.} BUI~\cite{Rodrigues2018Mendeley} is a breast cancer dataset collected with linear US probes. In this task, we need to classify whether the breast tumor that an image contains is malignant or not. This dataset contains 250 breast cancer images, where 100 samples are benign, and the rest 150 are malignant. We use the metrics of binary classification (\eg AUC) to evaluate the transferability of pre-trained models. The 5-fold cross-validation results are shown in Tab. \ref{tab:bui_compare}. In the BUI dataset, there are fewer images than in the POCUS, thus it needs better visual representations to train DNNs for gaining generalization on this task. In addition, videos in US-4 are derived with convex probes. The domain gap between different kinds of probes doesn't weaken the performance of Meta-USCL. Our results show that Meta-USCL reaches a similar performance improvement on BUI as POCUS (\eg its accuracy improvements compared with training from scratch are 10.3\% and 10.2\%). We hypothesize that the general US representations learned by Meta-USCL lead to a more powerful adaptation ability on a small US dataset than other methods. Therefore, it makes up for the performance loss brought by the domain gap. This hypothesis is validated in the following Sec. \ref{sect:analysis}. It is noteworthy that the BUI contains images with an average size of $120\times 90$, making the pre-training and fine-tuning with smaller inputs even more stable and powerful (Tab. \ref{tab:img_size_bui}). This is because when pre-trained on small images of US-4, the learned basic pattern distribution is similar to the BUI dataset.

\textbf{UDIAT-B Breast Tumor Segmentation.} The UDIAT-B~\cite{2017Automated} consists of 163 linear US probe breast tumor images from different women with a mean image size of $760\times570$ pixels. Each images presents one lesion with pixel-wise mask annotation. We choose this dataset to evaluate the model transferability for the downstream segmentation task. Of the 163 images, 53 are cancerous masses, and the other 110 are benign lesions. We randomly select 113 images for training and the other 50 images for validation. Since the segmentation task needs precise discrimination for pixels, we compare different methods with $224\times224$ pre-training image size for better detail resolving ability.

For segmentation, our main concern is the true positive (TP) region, which is the overlapped region between the ground-truth mask and the prediction. The importance of TP in clinical scenarios is primarily due to the low tolerance of missing any important lesion. Therefore, the three metrics $Dice=2TP/(FP+2TP+FN)$, $PPV=TP/(TP+FP)$, and $Sensitivity=TP/(TP+FN)$ are chosen to evaluate the segmentation performance, where TN, FP, FN are the number of true negative, false positive ,and false negative pixels. The final results are shown in Tab. \ref{tab:udiat_compare}. We can see that the Meta-USCL outperforms all other counterparts in this pixel-level prediction task. Note that both domain (linear/convex probe) and the task (segmentation/instance discrimination) between the downstream task and pre-training task are different, and the number of samples is also limited. The medical pre-training methods with contrastive image operations, including Meta+SimCLR and Meta-USCL, show the best performance compared with other supervised or unsupervised methods. This phenomenon indicates that: 1) instance-wise discrimination learns more pixel-level prediction ability than class-wise discrimination due to the semantic-invariant appearance discrepancies; 2) Meta-USCL reaches a significantly higher performance (3.1\%, $p<0.05$) compared to ImageNet, showing the potential for specialized representation learning methods to beat general pre-trained models on segmentation applications. 

\begin{figure*}[t]                 
\centering\centerline{\includegraphics[width=1.0\linewidth]{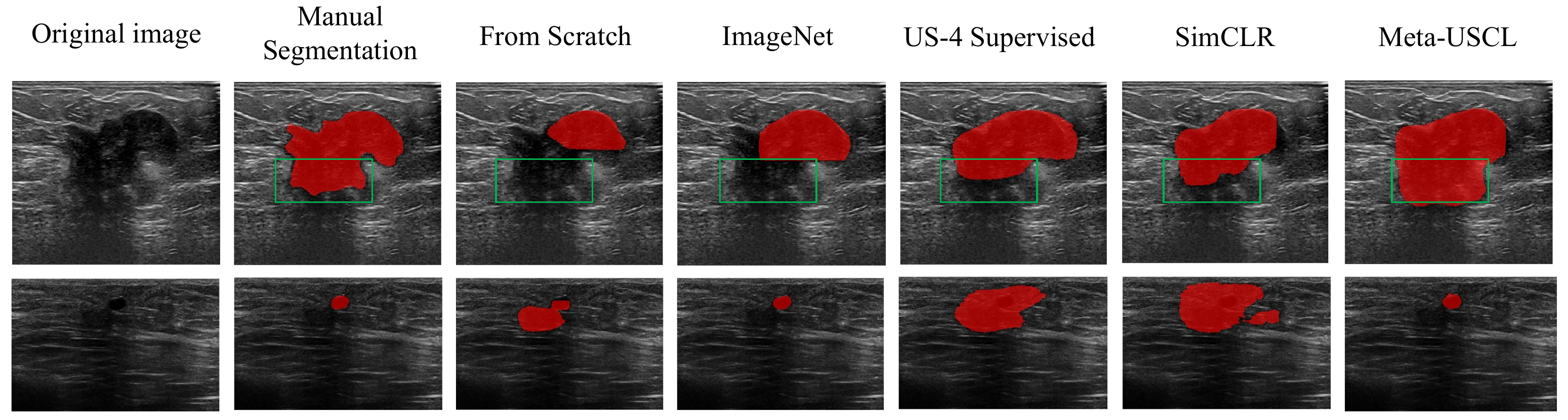}}
\caption{Examples of segmentation obtained with different pre-training methods. Meta-USCL pretrained models can both segment bigger tumors with precise edge, and localize smaller tumors accurately.}
\label{fig:seg}
\end{figure*}

In addition, we visualize the segmentation results in Fig. \ref{fig:seg}. The first row illustrates that Meta-USCL pre-trained model can distinguish the edge of a large tumor well. The upper half of this lesion is quite clear, resulting in consistent segmentation results of all methods. However, the lower half (see green rectangles in Fig. \ref{fig:seg}) becomes blurred, so better semantic understandings of the tumors are needed, where Meta-USCL performs the best. The second row shows that the Meta-USCL pre-trained model can localize a small lesion region precisely. Given an inconspicuous tumor, its outline is hard to distinguish from the black background area, making the detection of small lesions even more challenging than segmenting the edges of large lesions. Only ImageNet pre-training and Meta-USCL pre-training work well in this case.

\begin{figure*}[t]                
\centering\centerline{\includegraphics[width=1.0\linewidth]{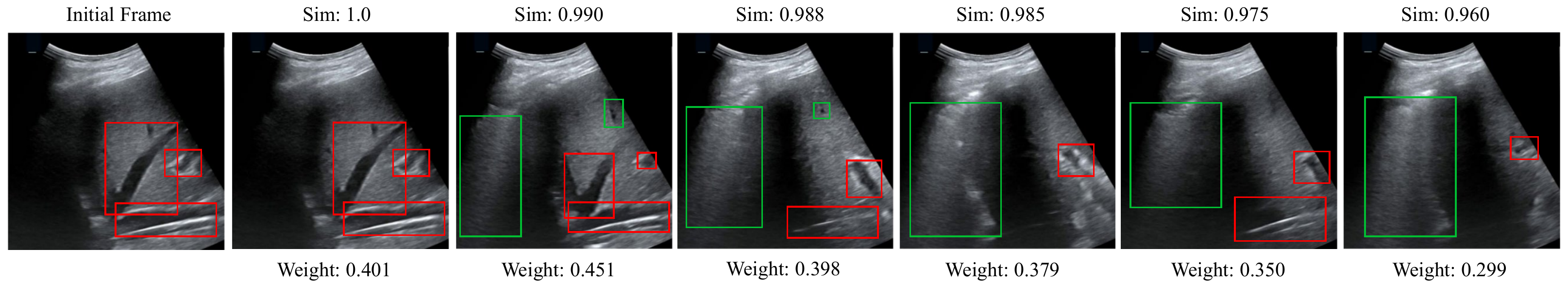}}
\caption{The illustration of weights and similarities between an initial frame and other frames. Red boxes highlight the changing semantic marks (\eg hepatic vein, portal vein and inferior vena cava) from the initial frame, green boxes denote the textural dissimilarities (gas in the lungs and a small portal vein section) in the following frames.}
\label{fig:sim1}
\end{figure*}

\subsection{Property Analysis and Visualization}\label{sect:analysis}

This subsection first explores the hyper-parameter tuning procedure. And then, how and why Meta-USCL works well is discussed through visualization of the knowledge that CMW-Net and the main model learn. Finally, we use quantitative analysis to reveal the performance \wrt the data amount of pre-training and fine-tuning.

\textbf{Hyper-parameter Tuning.} Previous works have shown that hyper-parameters are critical for contrastive learning to achieve desirable performance~\cite{Ting2020CVPR,he2020momentum}. Proper data augmentation and batch size are two of the most important factors that highly affect the transferabilities of learned image representations. Following SimCLR~\cite{Ting2020CVPR}, we try image cropping, color jittering, and Gaussian blurring in our experiments. The results are shown in Tab. \ref{tab:augment_tuning}. For both image cropping and color distortion, there are optimal parameters of moderate strengths. However, the Gaussian blurring does not show any performance gain for ultrasound contrastive learning. We conjecture that the imaging sharpness of the US systems is lower than the regular optical lens. The blurring leads to more unsatisfying image quality for US images than natural images. Therefore, Gaussian blurring is less helpful for models to learn transformation invariants since it cannot guarantee semantic consistency. For SimCLR, it was proved that the larger batch size improves the performance to a larger extent. But this conclusion does not hold for Meta-USCL (see Tab. \ref{tab:batchsize_tuning}). The batch size of 32 is the best for Meta-USCL, and the larger batch size even brings an evident drop in transfer learning performance. This phenomenon may be caused by the high similarity of frames from different US videos. Compared with natural images, US videos are far more similar to each other. Larger batch size increases the probability of sampling similar videos from the pre-training dataset, which may lead to the negative-pair similarity conflict problem.

\begin{table}[]
\centering
\caption{Tuning of data augmentation. The downstream performances are validated on the POCUS dataset. The cropping coefficient means the minimum random cropping ratio of the original image area. The color jittering coefficient means the maximum transform strength for brightness, contrast, and saturation.}
\renewcommand{\arraystretch}{1.0}
\small
\begin{tabular}{cccccc}
\toprule
Cropping & 0.6 & 0.7 & 0.8 & 0.85 & 0.9 \\
Acc (\%) & 92.3 & 93.5 & 94.1 & \textbf{94.6} & 94.3  \\
\midrule
Color jittering & 0.4 & 0.5 & 0.6 & 0.7 & 0.8 \\
Acc (\%) & 92.8 & 94.3 & \textbf{94.6} & 93.8 & 92.7  \\
\midrule
Gaussian blurring & & Yes & & No & \\
Acc (\%) & & 93.8 & & \textbf{94.6} & \\
\bottomrule
\end{tabular}
\label{tab:augment_tuning}
\end{table}

\begin{table}[]
\centering
\caption{Impact of batch size. Fine-tuning results on the POCUS show that larger batch sizes do not necessarily bring higher transfer performance.}
\renewcommand{\arraystretch}{1.0}
\small
\begin{tabular}{cccccc}
\toprule
Batch size & 8 & 16 & 32 & 64 & 128 \\
\midrule
Acc (\%) & 87.5 & 89.5 & \textbf{94.6} & 91.6 & 89.7  \\
\bottomrule
\end{tabular}
\label{tab:batchsize_tuning}
\end{table}

\textbf{Visualization of CMW-Net Learned Weights.} We visualize the sample pairs and the corresponding weights to verify that the meta-knowledge learned by CMW-Net can judge whether a positive pair is beneficial or not.
As shown in Fig. \ref{fig:sim1}, we use frames of a lung US video without any augmentation to obtain weights of sample pairs with CMW-Net. The leftmost image is the first frame of the video, and the following frames present descending cosine similarities to the first one. The resulting weights between the initial frame and the rest show a trend of rising and falling \wrt image similarities because the semantic marks are disappearing and the textural dissimilarities are becoming larger. The ideal sample pairs should have small semantic distances and appropriate texture information gaps between the two samples, which turns out to be moderately inconsistent (\eg the third frame and the initial frame in Fig. \ref{fig:sim1}).

\begin{figure}[t]              
\centering\centerline{\includegraphics[width=1.0\linewidth]{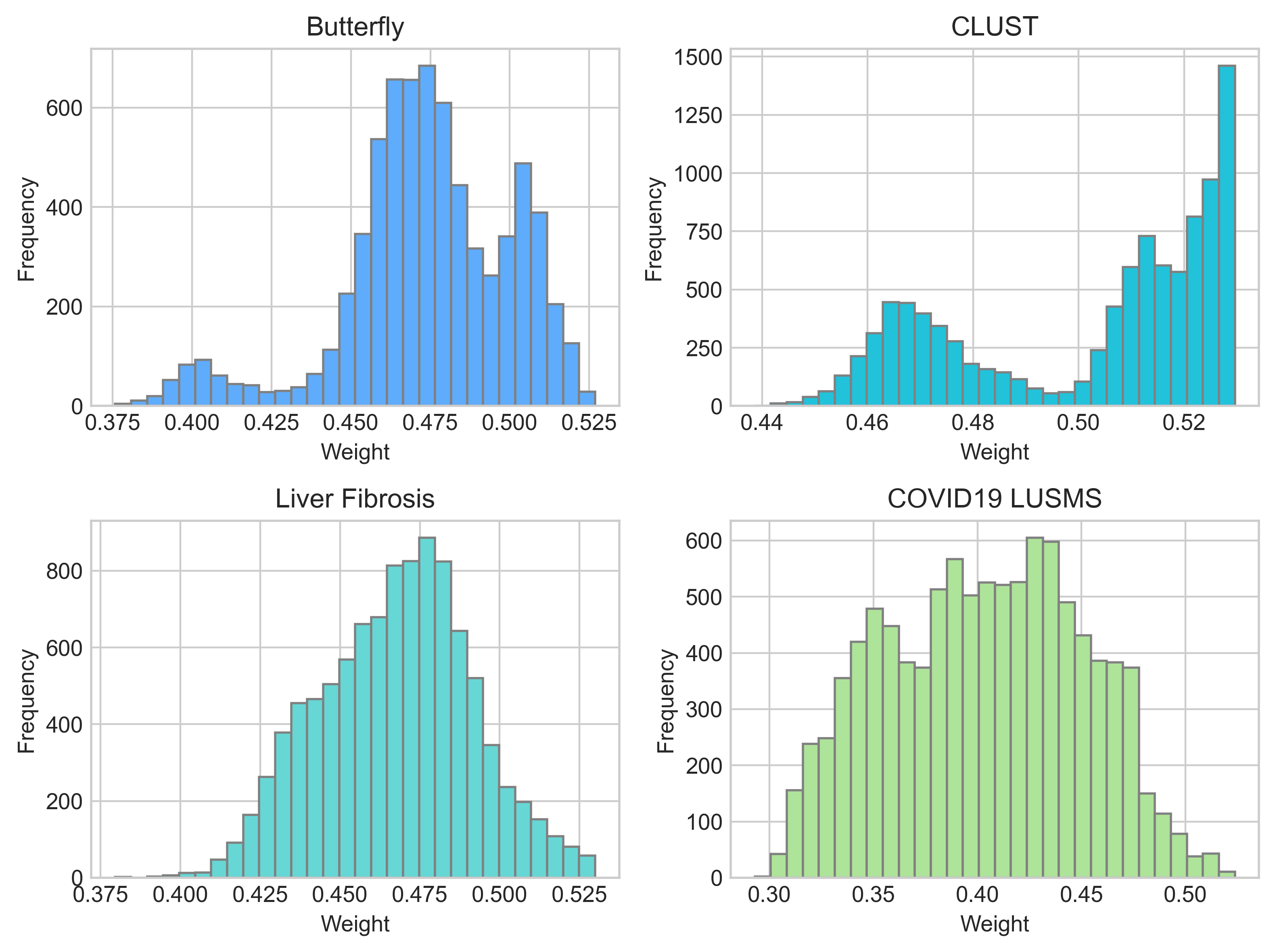}}
\caption{The weight distribution in each sub-dataset.}
\label{fig:hist_weight}
\end{figure}

\begin{figure}[t]                        
\centering\centerline{\includegraphics[width=1.0\linewidth]{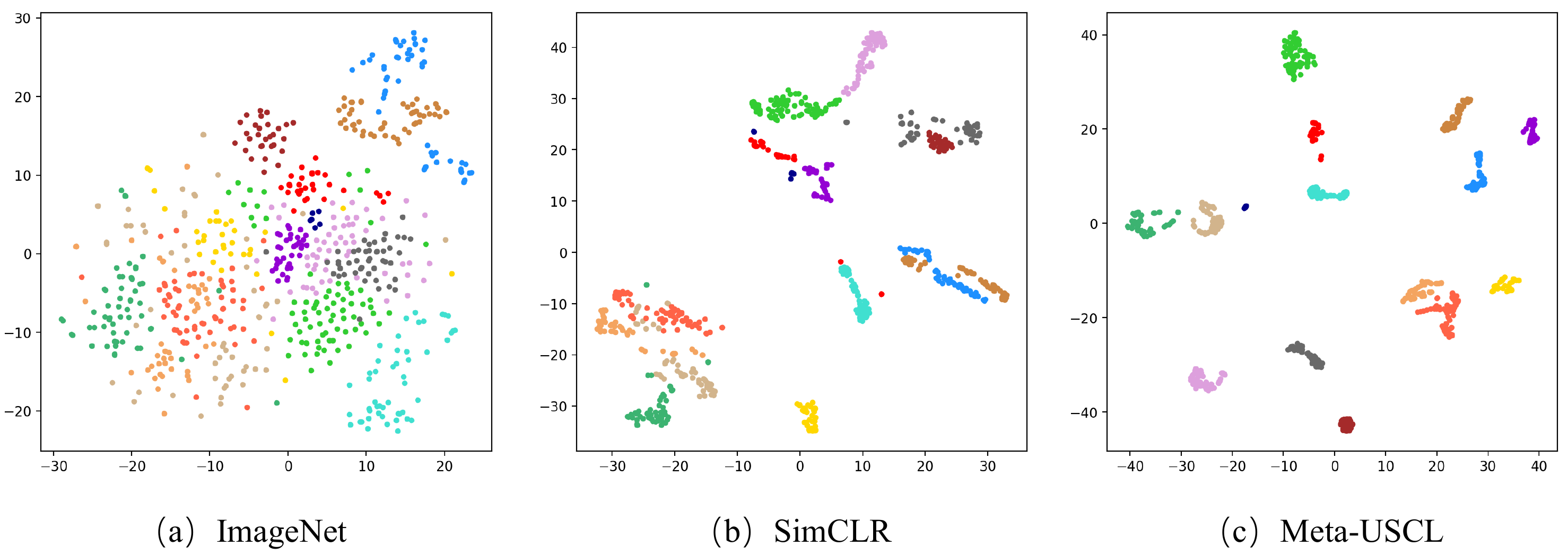}}
\caption{The feature spaces visualized by tSNE.}
\label{fig:tsne}
\end{figure}

In addition to different weights between positive pairs from a video, we also observe different weight distributions between different sub-datasets of US-4 (Fig. \ref{fig:hist_weight}), indicating that different US domains do not contribute their medical visual representations equally. For example, about half the sample pairs of the \textit{COVID19-LUSMS} sub-dataset have weights below 0.4, meaning they may be less important for pre-training. We infer that the different distributions are caused by the CMW-Net's target, minimizing the overall validation loss. The imbalanced US videos of different organs may make some patterns (\ie lesions) faster to be recognized, while the minority patterns are less focused. This results in the validation sample pairs of minority classes having higher loss values. For example, the \textit{COVID19-LUSMS} sub-dataset has the most abundant videos among the four sub-datasets. Therefore, CMW-Net assigns lower weights to it so the model can learn more from smaller subsets to reduce the overall validation loss. It is also noticeable that the differences between sub-datasets do not exactly match the number of videos per subset. We conjecture that the meta-reweighting network does not only consider the number of videos of different subsets. It is also affected by other factors such as the organ and lesion distributions, video lengths, and frame similarities (\eg the frame similarity of a CLUST video is higher than that of other sub-datasets, which leads to a lower probability of positive-pair dissimilarity defect and may be a reason for its higher overall weights).

All in all, CMW-Net makes contrastive learning work better by 1) emphasizing more on the sample pairs that help the model learn better semantic consistency; 2) balancing the sample pairs from different sample sub-domains to decrease the overall validation loss.

\textbf{Visualization of Representations Learned by Meta-USCL.} 
We use tSNE to visualize the feature space of the ImageNet, SimCLR, and Meta-USCL pre-training. Fig. \ref{fig:tsne} shows the feature distribution of randomly selected 15 US videos in the frame level. The ImageNet pre-trained DNNs are insensitive to US textures, resulting in scattered feature vectors. The reason may be that ImageNet pre-training does not involve sufficient US-related images, leading to the poor discriminative ability of US data. Using SimCLR to learn representations on US-4 can mitigate this issue. But the similarity defect problem makes the features from a video distributed on a broader range than Meta-USCL. This may be because SimCLR pushes two similar frames apart when they are sampled as a negative pair. However, Meta-USCL pre-training takes advantage of the characteristics of US videos to learn better feature representations, making all representations from a video distributed in a cluster. Overall, the performance advantage of Meta-USCL is due to two factors: (1) the pre-adaptation to the US domain and (2) its reasonable positive pair generating and weighting scheme.

\begin{figure}[t]
\centering\centerline{\includegraphics[width=1.0\linewidth]{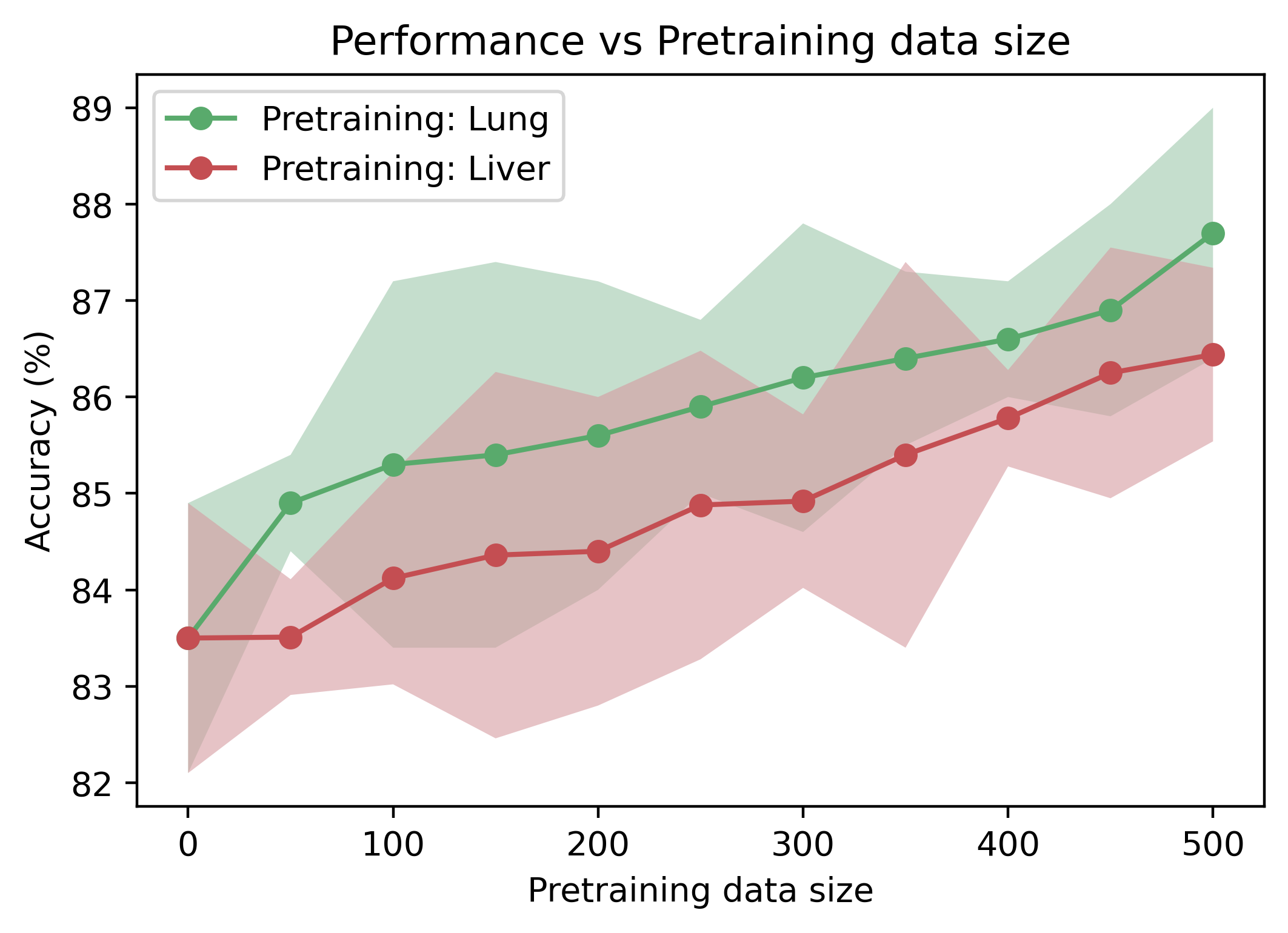}}
\caption{The relation between fine-tuning accuracy and pre-training data size. The shaded parts shows the standard deviations.}
\label{fig:acc_vs_datasize}
\end{figure}

\textbf{The Relationship between Pre-training Data Amount and Transfer Learning Performance.}
Unlike ImageNet, US-4 is a video dataset of US containing only 1300 videos. Therefore, it is very likely that the increasing amount of pre-training data can improve the pretraining performance. As illustrated in Fig. \ref{fig:acc_vs_datasize}, we only use one sub-dataset (\ie \textit{COVID19-LUSMS} or \textit{Liver Fibrosis}, which both have above 600 videos) to investigate the pre-training performance \wrt data amount and data domain. The pre-trained models consistently perform better on the POCUS pneumonia detection task with increased video amount.
Moreover, we find that the uptrend seems far from saturated when the video amount reaches 500, which means the larger US video dataset can further promote the pre-training performance.
In addition, the performance curve of lung pre-training is nearly always higher than liver pre-training, which might suggest that: 1) pre-training with Meta-USCL on the same domain gains more benefits than on a different domain; 2) pre-training in a different domain can also boost the performance of downstream tasks, though it is not an ideal solution.

\begin{table}[]
\centering
\caption{Fine-tuning accuracy (\%) on POCUS with different fine-tuning data size.}
\begin{tabular}{ccc}
\toprule
\multirow{2}{*}{Video ratio} & \multirow{2}{*}{ImageNet} & \multirow{2}{*}{Meta-USCL} \\
& & \\ \midrule
10\% & $66.7\pm1.8$ & $81.9\pm1.1$ \\
20\% & $68.9\pm1.6$ & $83.8\pm1.3$ \\
30\% & $71.4\pm1.9$ & $85.6\pm0.8$ \\
40\% & $73.3\pm1.2$ & $87.6\pm0.4$ \\
50\% & $75.3\pm0.8$ & $89.5\pm0.8$ \\
60\% & $77.6\pm1.3$ & $91.9\pm0.4$ \\
70\% & $80.1\pm0.7$ & $92.6\pm0.6$ \\
80\% & $83.1\pm0.9$ & $92.8\pm0.8$ \\
90\% & $83.6\pm0.5$ & $93.9\pm0.4$ \\
100\% & $84.2\pm0.8$ & $94.6\pm0.4$ \\ \bottomrule
\end{tabular}
\label{tab:acc_vs_dataratio}
\end{table}

\textbf{The Relationship between Fine-tuning Data Amount and Transfer Performance.} 
It was shown in Sec. \ref{sect: comparison} that Meta-USCL pre-trained models gain comparative or even greater advantages when fine-tuned on smaller datasets than ImageNet pre-training. We demonstrate fine-tuning results with different video ratios of the POCUS dataset in Tab. \ref{tab:acc_vs_dataratio}.
As the data ratio decreases from 100\% to 10\%, the ImageNet pre-trained model has a 17.5\% performance drop, but Meta-USCL only shows a 12.7\% difference. The result indicates that when fine-tuning on a small amount of data, Meta-USCL pre-trained models generalize the pattern of downstream tasks better since they are pre-trained on the US video domains themselves. The network parameters of these pre-trained networks can be regarded as closer to optimal points of different downstream tasks than ImageNet pre-training, leading to easier generalization.

\section{Conclusion}

In this work, we propose a novel visual representation learning method, Meta-USCL, especially for US downstream tasks. Meta-USCL generates sample pairs with the PPI module to learn rich feature representations from the limited US video datasets. Moreover, it uses CMW-Net to choose positive sample pairs with more visual contrastive information. We conduct extensive experiments to validate the superiority of Meta-USCL to other existing representation learning methods, including ImageNet pre-training. Its outstanding performance on small-size target datasets makes us envision that Meta-USCL may serve as a primary source of transfer learning for US applications. Therefore, we make the Meta-USCL framework an open resource and encourage researchers all around the field to contribute to its improvements and further explorations.

\bibliographystyle{IEEEtran}
\bibliography{MetaUSCL}

\end{document}